# Robustness-aware 2-bit quantization with real-time performance for neural network[1]

Xiaobin Li[1], Hongxu Jiang[1,*], Shuangxi Huang[1] and Fangzheng Tian[1]

[1] Beijing Key Laboratory of Digital Media, Beihang University, Beijing 100191, China

**Abstract.** Quantized neural network (NN) with a reduced bit precision is an effective solution to reduces the computational and memory resource requirements and plays a vital role in machine learning. However, it is still challenging to avoid the significant accuracy degradation due to its numerical approximation and lower redundancy. In this paper, a novel robustness-aware 2-bit quantization scheme is proposed for NN base on binary NN and generative adversarial network(GAN), witch improves the performance by enriching the information of binary NN, efficiently extract the structural information and considering the robustness of the quantized NN. Specifically, using shift addition operation to replace the multiply-accumulate in the quantization process witch can effectively speed the NN. Meanwhile, a structural loss between the original NN and quantized NN is proposed to such that the structural information of data is preserved after quantization. The structural information learned from NN not only plays an important role in improving the performance but also allows for further fine tuning of the quantization network by applying the Lipschitz constraint to the structural loss. In addition, we also for the first time take the robustness of the quantized NN into consideration and propose a non-sensitive perturbation loss function by introducing an extraneous term of spectral norm. The experiments are conducted on CIFAR-10 and ImageNet datasets with popular NN( such as MoblieNetV2, SqueezeNet, ResNet20, etc). The experimental results show that the proposed algorithm is more competitive under 2-bit-precision than the state-of-the-art quantization methods. Meanwhile, the experimental results also demonstrate that the proposed method is robust under the FGSM adversarial samples attack.

## 1    Introduction

Deep learning methods have been successfully applied to different computer vision and multimedia tasks with the introduction of many advanced DNN(deep neural network) architectures, such as ResNet50 and VGG16. Although these DNNs show excellent performances in many multimedia applications, the complicated network structure and huge number of parameters hinder their applications in embedded platforms with limited computation and storage resources.

Low-bit quantization as a key technology of deep compression can not only relief the storage issue, but also improve the computing efficiency in the meantime, which is friendly in embedded equipment. Although several low-bit presentation NNs (e.g. Binary neural network[1], XNOT network[2] and ternary network[3]) have shown significant advantages on computation and storage resources, their performance drop significantly compared to the unquantized counterparts. The recently proposed quantization methods INQ[4] and balanced quantization[5] achieve both efficient weights quantization and high precision. Furthermore, DoReFa-NET[6] and PACT[7] propose to quantize both the convolutional layers and activation layers, which achieve the acceleration of neural networks in real sense and obtain higher accuracy than XNOT network, and thus are widely used in industry. Although the previous algorithms may work well on various DNNs, mostly are adopt multiply-accumulate in the quantization process. Even with the low bit precision, the calculation process has not been accelerated, which can still lead to potential computational burden on the embedded platforms and fail to achieve real-time performance. Though binary NN with 1-bit precision and shift addition process, but its information loss with lead to poor precision. To this end, quantized NN with shift addition operation is an urgent need in some extreme resource limited scenarios, such as auto pilot and spaceflight.

* Corresponding author
E-mail addresses: lixiaobin@buaa.edu.cn (X. Li), jianghx@buaa.edu.cn (H. Jiang), huangsx@buaa.edu.cn (S. Huang), amazingtian@buaa.edu.cn (F. Tian).

In addition, we observe that the robustness of the NN might also be affected after quantization. However, to the best of our knowledge, no previous work considers the robustness of the quantized NN and none of them evaluate the performance on the adversarial examples. To this end, a novel robustness-aware 2-bit quantization scheme for NN is proposed, which has better accuracy and robustness than vanilla quantization algorithms. Specifically, we improve the accuracy of the quantized LNN by taking the structure information into consideration, which is inspired by NAS(network architecture search). To improve the robustness of the quantized LNN, we propose to use loss sensitive generative adversarial network (LS-GAN) with a lipschitz constraint, which is known for improving the robustness of neural networks. Morevover, we add spectral norm sub-item as the non-sensitive perturbation loss to further constrain the network.

The contributions of this paper are summarized as follows:

- A novel 2-bit quantization scheme is proposed, which focuses on three major bottlenecks of embedded applications of traditional NN: 1. limited storage capability; 2. hardware unfriendly; 3.real-time performance.
- In order to improve the robustness and accuracy of NN, we employed LS-GAN(loss-sensitive GAN)[8] and non-sensitive perturbation loss in our framework. Meanwhile, we design the specific quantization dictionary to realize NN's accelerating.
- To further exploit the variations in the data and improve the performance of the quantized NN, the structural information is preserved in the quantized NN. Note that the self-reference quantization module can also be readily plugged into other quantization methods, which also testifies its effectiveness on current state-of-the-art quantization algorithms, e.g. DoReFa and PACT.
- Extensive experiments have verified the effectiveness of the proposed framework, with respect to both accuracy and robustness.

## 2 Related work

Our work is most related to binary NN, structural knowledge distillation and self-reference learning. Therefore, we discuss the recent works in these fields in this section.

### 2.1 Binary NN

Network binarization beginning with BNN and XNOR-Net aims to accelerate the inference of neural networks and save memory occupancy without much accuracy degradation. As we know, shift operation is very hardware friendly and the most efficient approach to speed up low-precision networks. By directly binarizing the 32-bit parameters in DNNs including weights and activations, we can achieve significant accelerations and memory reductions. Meanwhile, network binarization usually accompanied by large degree of accuracy loss. The existing optimization method can be summarized as minimized quantization error, improved loss function and reduced gradient error. WRPN [24] and Shen et al. [25] improve the accuracy of BNN by increasing the number of channels in the convolutional layer, and not only achieved accuracy improving but make BNN wider and increase the computational complexity to a large extent, simultaneously. TWN and TTQ[19] enhance the representation ability of neural networks with more available quantization points. ABC-Net[20] recommends using more binary bases for weights and activations to improve accuracy, while compression and acceleration ratios are reduced accordingly. The proposed HWGQ[21] considering the quantization error from the aspect of activation function. Further proposed LQ-Net[22] with more training parameters, which achieved comparable results on the ImageNet benchmark but increased the memory overhead. IR-Net[23] take information loss into account and introduced information entropy to the loss function, witch has a better accuracy performance.

Although much researches has been made on network binary networks, the existing quantization methods still cause a significant drop of accuracy compared with the full-precision models. Compared with other model compression methods, such as pruning and matrix decomposition, network binarization can greatly reduce the memory consumption of the model, and make the model fully compatible with shift operations to get good acceleration.

## 2.2 Knowledge distillation

Knowledge distillation is a technique to transfer knowledge from a complex network (teacher network) to a compact network (student network), including data knowledge distillation and structure knowledge distillation. Wu et al. train the student network by taking the output of the teacher network as the soft target, and achieve the knowledge transfer by replacing the L2 loss with cross entropy loss[14]. Adriana Romero et al. fit the complexity of the teacher network by inputting more no-tag data into the student network. Junho Yim et al. optimized the knowledge transfer by refining knowledge distillation to the layer [15]. For the first time, we use knowledge distillation to extract the structural information of the network and construct new loss function to guide the quantization process of the network.

## 2.3 Self-reference learning

In the early 1940s, John von Neumann has raised the concept of an artificial self-replicating machine, which is prior to the discovery of DNA's role as the physical mechanism for biological replication. Specifically, Von Neumann demonstrates a configuration of initial states and transformation rules for a cellular automaton that produces copies of the initial cell states after running for a fixed number of steps.

Self-reference learning can be seen as one special case of adversarial learning. Adversarial learning is a technology that advances each other through the adversarial training process. GAN is a typical technology of adversarial learning, which is widely used in text production and image synthesis. For example, GAN can be used in many tasks, such as image style migration, image generation, image coloring. In addition, the constrained CGAN can be used for the conversion of text-to-image. The idea of adversarial learning is also employed in pose estimation [26], encouraging the human pose estimation result not to be distinguished from the ground-truth. Semantic segmentation encourage the estimated segmentation map not to be distinguished from the ground-truth map[27]. However, in the depth prediction task, the ground truth maps are not discrete labels and further method [28] use the ground truth maps as the real samples. Different from theirs, our method keep the structure of the student network and the teacher network in sync, aim to transfer the structure information to the student networks and realize the true sense of self-learning.

## 3 2-bit incremental quantization

Binary neural networks have received widespread attention from industrial community because of their small storage capacity and high inference efficiency. However, compared with the full-precision corresponding method, the accuracy of the existing quantization method still has a significant drop. There are two major factors of drop: information loss during forward propagation and inaccurate of gradient information in back propagation. In order to compensate for the loss of network information caused by binarization, this paper is based on the evolution of an excellent binary network to a 2bit quantitative model that can meet the actual project accuracy and real-time requirements.

## 3.1 Preprocessing of parameter distribution

The research on deep compression tells us that the redundancy of deep neural networks is very large, large weights matter more than small ones. In order to optimize the expression ability of neural networks, we expect increases the proportion of large weights by changing the distribution of model weights. Therefore, in this section, we takes AlexNet and ResNet-18 networks as examples to analyze the initial distribution of network model parameters, as shown in the figure 1.

It can be seen from Fig.1. that the overall weight distribution trend is close to the normal distribution, but in actual projects, σ (total variance) is often unknown, and s (sample variance) is often used as the estimated value of σ, and the weight distribution of different networks is slightly different. In order to fit the initial distribution of the network as much as possible and retain more network information, we introduced the T distribution. T-distribution conversion can make the distribution of weights more uniform, try to increase the probability density of large weights, and find key weights.

However, for neural networks, the quantization is equal to the parametric sampling process. We hope to fit the internal full-precision parameter with fewer parameter samples. According to the law of large numbers and central-limit theorem in mathematical statistics: when the sample size is large enough, the mean value of these samples is infinitely close to the expectation of the population, and no matter what distribution the sample population obeys, the sample mean fluctuates around the population mean as a normal distribution[29-30]. Therefore, it also proves that the lager parameters scale the closer to the normal distribution.

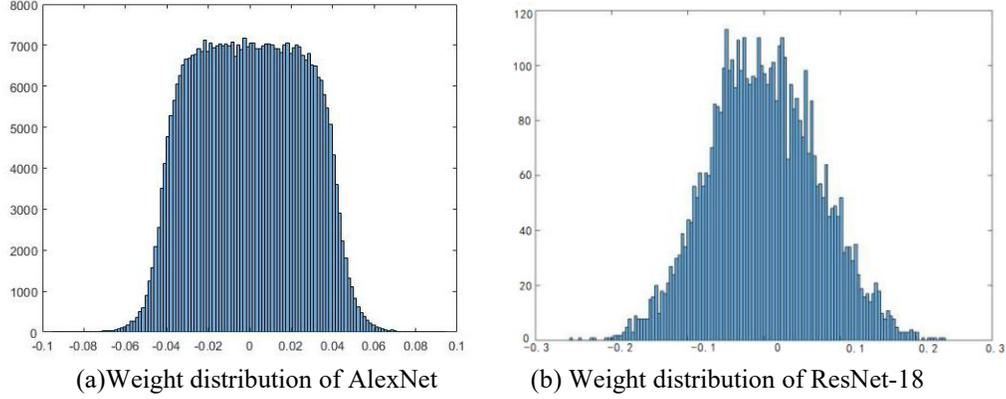

(a)Weight distribution of AlexNet　　　　(b) Weight distribution of ResNet-18

**Fig. 1.** Network weight distribution of Alexnet and ResNet-18

The probability density function of the T distribution is shown as follows,

$$f(x;n) = \frac{\Gamma\left(\frac{n+1}{2}\right)}{\sqrt{n\pi}\,\Gamma\left(\frac{n}{2}\right)}\left(1+\frac{x^2}{n}\right)^{-\frac{n+1}{2}}, -\infty < x < +\infty \tag{1}$$

Where $x$ indicates sample data, $n$ indicates degree of freedom for the T-distribution. And when $n \to \infty$, the T-distribution is going to be normal distribution, show as the formula(2).

$$f(x;n) = \frac{1}{\sqrt{2\pi}} e^{-\frac{x^2}{2}}, -\infty < x < +\infty \tag{2}$$

In this paper, the quantization process is associated with the degree of freedom of T-distribution, and we obtain more accurate quantization values through reducing the T-distribution' degree of freedom. Meanwhile, as shown in Fig.2, reducing the degree of freedom($n$) will increase the density of the large parameters and improve the representation ability of the neural networks. Then, we can optimize the pre-training model by adjusting the value of $n$.

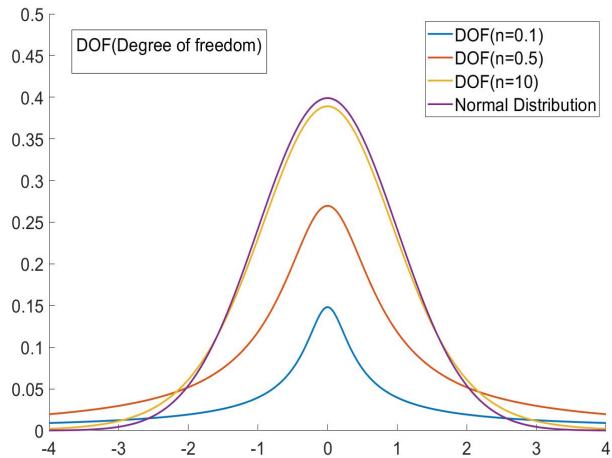

**Fig. 2.** Probability density of the T distribution

## 3.1 Hardware friendly 2-bit quantization

For hardware devices, the most friendly calculate method is shift-add operation, which also brings the birth of binary network, and greatly reduces the storage of neural networks and improves the computing efficiency. Based on binary network and inspired by IR-Net[23], this section designs a hardware friendly 2-bit quantization method. The advantages of this method are as follows: (A) all quantization parameters are represented by the combination of shift and plus, which greatly improves the operation speed of the model; (B) minimize the information loss during the forward propagation of the neural networks by introducing the information entropy into the object optimizer, and maximizing the information entropy to ensure the accuracy of the 2-bit quantization is state-of-arts. The specific quantization process is as follows:

In the forward propagation of neural network, the usual optimal quantizer by minimizing the quantization is show in the formula(3),

$$\min J(Q_x(x)) = \|x - Q_x(x)\|^2 \tag{3}$$

Where x indicates full-precision parameters, $Q_x(x)$ indicates quantization parameters. However, this optimizer ignores the information loss of the network in the forward propagation. The quantity of information can be description by information entropy( $H(Q_x(x))$ ), as shown in Formula (4),

$$H(Q_x(x)) = -p\ln(p) - (1-p)\ln(1-p) \tag{4}$$

Where $p$ and $1-p$ indicate the probability of binarization parameter, +1 and -1, respectively. Consequently, we build the new object optimal quantizer by formula(5), therein, $\lambda$ is the super parameter.

$$\min J(Q_x(x)) - \lambda H(Q_x(x)) \tag{5}$$

Because of the binarization parameter variables follow the Bernoulli distribution, under the Bernoulli distribution assumption, when $p = 0.5$, the information entropy of the quantized values takes the maximum value.

(b) Sampling parameters and standard the T-distribution, then, we can adjust the degree of freedom $n$ to enlarge the proportion of large number parameters.

(c) Retain the ability to update the backpropagation algorithm. Keep the derivative value of the gradient estimation function close to 1, and then gradually reduce the cutoff value from a large number to 1, thereby ensuring the ability to update early in training. The progress shown as Fig.3.

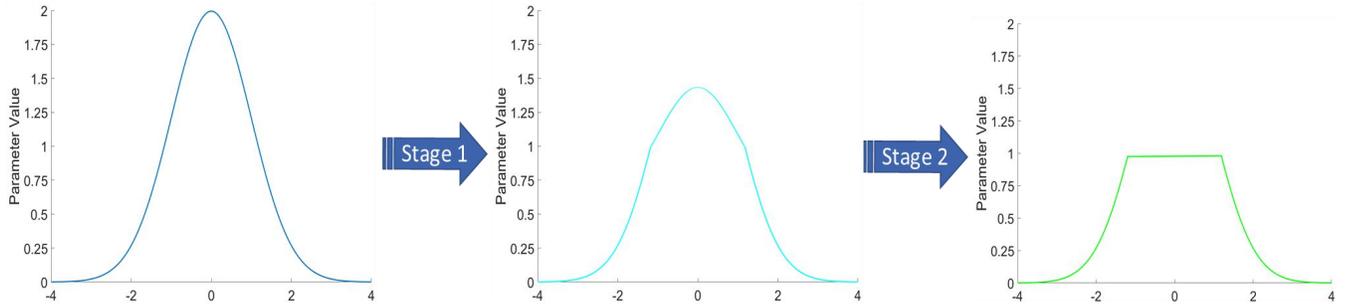

**Fig. 3.** Gradually reduce progress of the cutoff value

(d) The parameters near zero are updated more accurately. For the parameters that have been updated in the fixation process (c), calculate the coordinates (x, y) of the inflection point of the T distribution, keep the truncation as x, and gradually evolve the derivative curve to the shape of the step function to complete all network parameter updates.

We set $(f(x:n))''(x) = 0$, thus obtained $x_1$ and $x_2$.

(d) The obtained value $x_1$ and $x_2$ is approximated to $\frac{1}{8}\text{INT}(s)$, where $s \in [1,7]$, then a 2-bit quantization dictionary [-1, $x_1$, $x_2$, 1] is obtained.

$$\frac{1}{8}\text{INT}(s) = 2^a + 2^b \tag{6}$$

Where $\{a,b\} \in [-3,-1]$, ultimately, we cluster the remaining parameters to the quantization dictionary and accomplish the 2-bit quantization.

Such quantization can satisfy all quantized values that can be decomposed into exponential power addition form of 2, which is hardware friendly. The clock cycle required for shift-add operation is about 1/4 of that of floating-point operation, so it can be realized 4 times the theoretical network acceleration effect.

**Algorithm 1 Hardware friendly 2 bit quantization**

1: **Build the object optimal quantizer**
   Set $p = 0.5$, calculate the maximum information entropy $H(Q_x(\mathrm{x}))$;
   Introduce $H(Q_x(\mathrm{x}))$ into formula(3) and build the new object optimal quantizer;
2: **T-distribution conversion**
   Adjust the degree of freedom $n$ to enlarge the proportion of large number parameters;
3: **Cutoff of the parameters**
   When parameter $|x| > 1$;
   $x = \mathrm{sgn}(x)$;
4: **Build the 2-bit quantization dictionary**
   Set $(f(x:n))''(x) = 0$;
   Calculate the $x_1$ and $x_2$;
   $\{x_1, x_2\} \approx \frac{1}{8} INT(S)$, where $s \in [1,7]$;
   Then, build the quantization dictionary $[-1, x_1, x_2, 1]$;
5: **Update the parameters of neural network**
   Calculate the gradient of weight(W) and activation(A) parameters;
   $W, A \rightarrow -1, 2^a + 2^b, +1$;

## 4 Robustness-aware quantization FOR NN

Based on the analysis in Section 3, this section presents our method on resolving the problems of accuracy degeneration and the reduction of robustness in the quantized NN.

In this paper, the lipschitz constrained LS-GAN is adopted to improve the accuracy and robustness of the NN, so that the lightweight NN has certain defense capabilities and is more friendly to practical applications. Inspired by the idea of NASnet, we need to pay more attention to structural information. Thus, we propose to impose the structural consistency between the quantized NN and the original NN, and thus the structural information can be used to guide the quantization process.

### 4.1 Self-reference quantization scheme

We first use GAN to optimize the progress of NN's quantization. Here, we choose to use LS-GAN since the generated sample density is proved to be consistent with the real density. In addition, we focus on handling the remote generated samples to the real samples, such that LS-GAN obtains "on-demand distribution" and avoid the problem of gradient vanishing. Furthermore, LS-GAN's weak constraint can improve the robustness to some extent. Therefore, we employed LS-GAN in our framework, which is named self-reference quantization. As shown as the Fig.4, the overall framework mainly contains three parts: original NN, quantized NN and discriminator. In the field of image processing, we always treat the deep neural network as a molectron which is composed of feature extractor and classifier. Therefore, we will get the real feature map $F_f$ and fake feature map $F_r$ using the feature extractors of original NN and quantized NN, respectively. Meanwhile, the feature extractor of quantized NN can be seen as the generator of LS-GAN.

Implementation details of self-reference quantization are shown as follows：
- Obtain the optimal pre-trained model as the original NN;
- Fix the original NN and randomly initialize the quantized NN;
- Sample one image form the dataset as the input of original NN and quantized NN, then get the corresponding $F_f$ and $F_r$

- Feed the $F_f$ and $F_r$ into discriminator and get a score. Update the discriminator in the case of $F_r$ get low score and $F_f$ get high score, otherwise update the generator by the total loss which is composed of non-sensitive perturbation loss *L(S)* and structural loss $L_\theta(G_\phi*)$ ;
- Obtain the optimal quantized NN by the equilibrium between discriminator and generator.

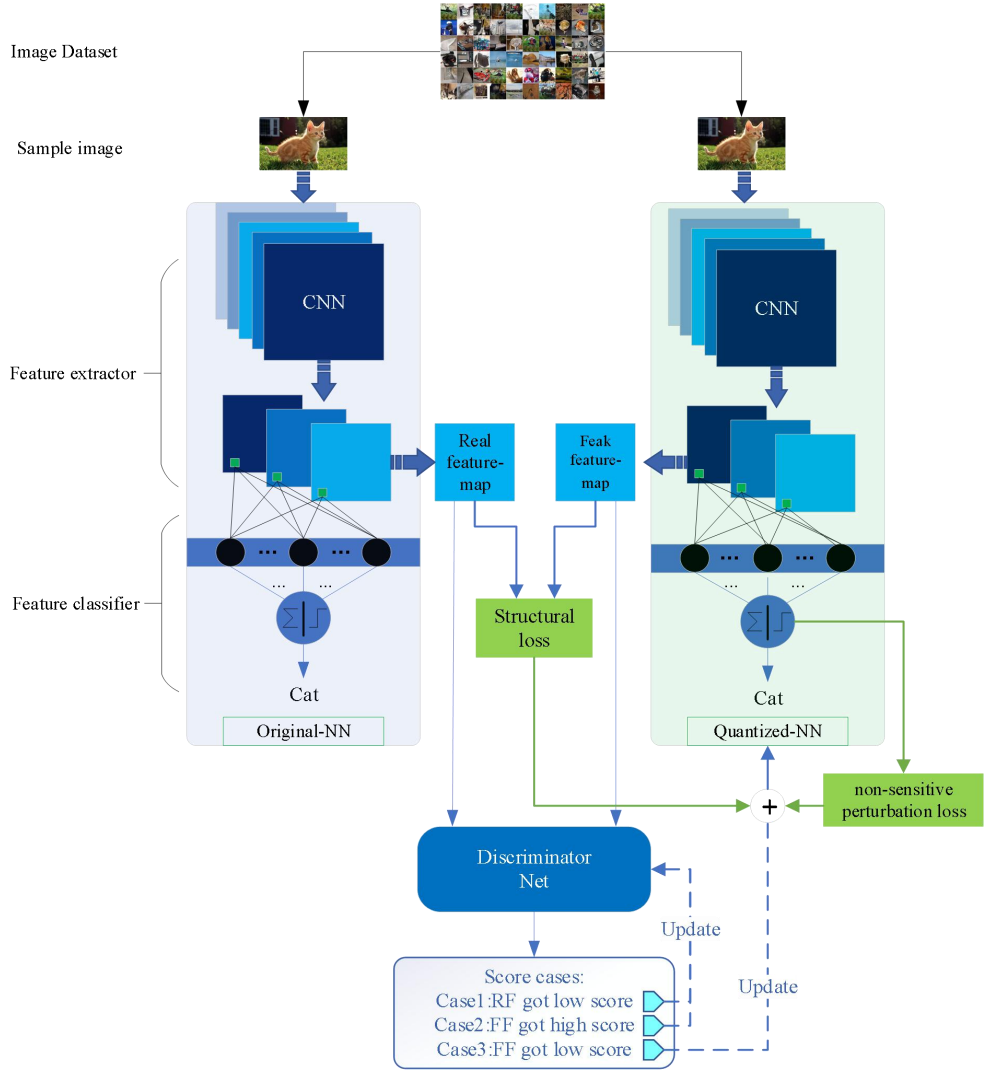

**Fig. 4.** Self-reference quantization scheme

As shown above, we use the pre-trained model as the original NN, which aims to let the generator learn the same structure information. Original NN and Quantized NN will have the similar high dimensional features and identical size of feature maps with same structure, also can speed and smooth the models' convergence procedure. Therefore we call our method self-reference quantization.

The feature maps obtained from the feature extractor contain hidden structural information of the data. Therefore, we expect that the $F_f$ and $F_r$ to be as similar as possible, such that the classification results will be more similar. The motivation is that although the original NN and quantized NN have the same network structure, the data characteristics will change greatly after quantization. In consequence, the importance of structural information should be considered. The quantized NN with our algorithm is invariant to small perturbation and has stronger robustness by means of adversarial training and the lipschitz constrained LS-GAN.

The discriminator is composed of two full-connection layers. Intuitively, the discriminator network should not be too complicated, otherwise the discriminant ability is too strong, leading to the sharply turbulence in the updating process and unstable results. We use the feature maps after feature extraction as the input of the discriminator, and get a score after the feature map pass the discriminator.

## 4.2 Non-sensitive perturbation loss

As we know, the Lipshitz is a weak constraint. Therefore, only using LS-GAN to improve the robustness of quantization models is insufficient. In this section, we add spectral norm sub-item as the non-sensitive perturbation loss to stronger the constraint.

For any network model $f_w(x)$, if the $f_w(x+\Delta x)$ output the same classification results after minor perturbations $\Delta x$, the model is said to be robust and can resist certain attacks. If a cat is identified as a dog by the neural network after adding some noise is not acceptable in engineering practice. We expect the network to be controlled by linear functions, so one of the most straightforward way is to introduce Lipschitz conditions to constrain the neural network.

We firstly present the non-sensitive perturbation loss *L(S)*, together with the Lipschitz constraint to make the model have stronger generalization ability and robustness by gathering the forefathers' achievement[18]. Specifically, the Spectral Norm Regularization is introduced into the loss, and the square of Spectral Norm is used as an additional regularization to replace the classical L2, which is formulated as follows:

$$L(S) = loss(y, f_w(x)) + \lambda \|W\|_2^2 \tag{7}$$

Where $loss(y, f_w(x))$ is the original QNN's loss, $W$ is the weight matrix, x and y denote the model's input and output. The formula shows that the loss terms and weight parameters constraining term work jointly, suggesting that we can constrain the parameters by the $\lambda \|W\|_2^2$. In fact, spectral norm is equivalent to the absolute value of $W^TW$'s maximum characteristic root in a neural network. So we only need to make sure the matrix $W$ satisfies the self-orthogonal condition. Eventually, the parameters can be updated by the *L(S)*.

## 4.3 Structural information mapping

Our method using LS-GAN to distill the structural information of teacher network and guide the student network training, then minimize the divergence of output feature maps between original NN and the quantized NN. Since our focus is the lightweight NN with smiple structure we only considered the last layer of feature extraction, which can capture and map the whole structural information into the quantized NN and accelerate the model convergence.

In this section, we calculate the divergence between $F_r$ and $F_f$ to obtain the structural difference between the original NN and quantized NN. We convert the image classification task to the classification of feature map, which equivalent to mapping the original data from high-dimensional space to low-dimensional space. We can learn a loss function $L_\theta(x)$ by assuming that a $F_r$ ought to have a smaller loss than a $F_f$ by a desired margin. Then the generator can be trained to generate realistic samples by minimizing their losses. Thus, the network can be trained to distinguish $F_r$ and $F_f$ with the following constraint between losses:

$$L_\theta(F_r) \leq L_\theta(F_f) - \Delta(F_r, F_f) \tag{8}$$

Where $\Delta(F_r, F_f)$ is the margin measuring the difference between $F_r$ and $F_f$. Because of the original NN is our fixed pretraining model, so the constraint question can transfer to a minimization problem by fixing the generator $G_\phi *$ and learning the $L_\theta(G_\phi *)$. Simultaneously, LS-GAN introduces a nonnegative slack variable $\xi_{F_r, F_f}$ to solve data invisible problem as follows,

$$\min_{\theta, \xi_{F_r, F_f}} \mathbb{E}_{F_r \sim P_{F_r}} L_\theta(F_r) + \lambda \mathbb{E}_{\substack{F_r \sim P_{F_r} \\ F_f \sim P_{F_f}}} \xi_{F_r, F_f}$$

$$s.t., L_\theta(F_r) - \xi_{F_r, F(F)} \leq L_\theta(G_\phi *) - \Delta(F_r, F_f) \qquad (9)$$

$$\xi_{F_r, F_f} \geq 0$$

Where $\lambda$ is a positive tradeoff parameter, $P_{F_r}$ and $P_{F_f}$ are the data distribution of $F_r$ and $F_f$. As long as we learned $L_\theta(G_\phi *)$, we can fix it and get the optimal $G_\phi *$ by solving the following minimization problem:

$$\min_\phi \mathbb{E}_{F_f \sim P_{F_f}} L_\theta(G_\phi *) \qquad (10)$$

The LS-GAN optimizes $L_\theta(G_\phi *)$ and $G_\phi *$ alternately by seeking an equilibrium ($L_\theta(G_\phi *)$, $G_\phi *$), and the $L_\theta(G_\phi *)$ contains structural knowledge of original NN we want. In the next section, we will combine it to train our quantized NN.

After obtaining the structural-information loss $L_\theta(G_\phi *)$, we make a comparison experiment on ImageNet dataset with other distillation loss, shown in Table 4(From the section 5.4).

## 5 Experiments

Our quantization scheme is implemented in Pytorch. The experiments are conducted on CIFAR-10 and ImageNet data sets. The results demonstrate the effectiveness of our method compared to the well-known quantization modules: DoReFa and PACT. In order to ensure the fair comparison, we conduct the experiments with the same hyper-parameters. All the models are trained using SGD for 250 epochs.

The experiments of quantized NN on CIFAR-10 and ImageNet are conducted with number of bits ranging from 2 to 4. Table 1 and Table 2 show the comparison results of SOTA method with our 2-bit quantization scheme on CIFAR-10 and ImageNet, respectively. Firstly, we report the accuracy of quantization method of DoReFa, PACT and LQ-net separately. Then, we take our scheme as an additional optimizer on these two outstanding quantization method and report the results. The weights and activations of all the models are quantized simultaneously with the same bit numbers.

Table 1. The performance of 2bit quantized ResNet20 and SqueezeNet on CIFAR-10 dataset.

(a) ResNet20 accuracy of 2-bit quantization

| Method | FP accuracy(%) | 2-bit-Quantization accuracy(%) | MAdd |
|---|---|---|---|
| DoReFa | 91.8 | 88.2 | 1330M |
| PACT | 91.8 | 89.4 | 1324M |
| LQ-net | 91.8 | 89.2 | 1310M |
| **Ours** | 91.8 | **89.8** | **795M** |

(b) SqueezeNet accuracy of 2-bit quantization

| Method | FP accuracy(%) | 2-bit-Quantization accuracy(%) | MAdd |
|---|---|---|---|
| DoReFa | 92.5 | 83.9 | 211M |
| PACT | 92.5 | 85.4 | 202M |
| LQ-net | 92.5 | 85.1 | 205M |
| **Ours** | 92.5 | **86.3** | **134M** |

Table 2. The performance of quantized ResNet18 and MoblieNetV2 on ImageNet dataset.

(a) ResNet18 accuracy of 2-bit quantization

| Method | FP accuracy(%) | 2-bit-Quantization accuracy(%) | MAdd |
|---|---|---|---|
| DoReFa | 70.4 | 62.6 | 3789M |
| PACT | 70.4 | 67.0 | 3740M |
| LQ-net | 70.4 | 66.1 | 3759M |
| **Ours** | 70.4 | **67.5** | **1922M** |

(b) MoblieNetV2 accuracy of 2-bit quantization

| Method | FP accuracy(%) | 2-bit-Quantization accuracy(%) | MAdd |
|---|---|---|---|
| DoReFa | 71.7 | 63.7 | 300M |
| PACT | 71.7 | 67.5 | 305M |
| LQ-net | 71.7 | 64.9 | 299M |
| **Ours** | 71.7 | **67.9** | **174M** |

From the Table 1 and Table 2, it can be observed that the accuracy of quantized SqueezeNet and MoblieNetV2 reduce seriously, by almost 8% accuracy drops. By contrast, our 2-bit quantization scheme improved the accuracy of other SOTA method to a large degree.

In order to verify the real-time performance of the algorithm, we simulated the first layer of ResNet50 under the hardware of Xilinx zcu102 with the clock cycle is 100M and the input image size is 224*224. We compared our 2-bit-quantization algorithm with the vanilla 2-bit-quantization algorithm in Table 3.

**Table 3.** The delay comparison of our 2-bit-quantization method with the vanilla.

| Method | 2-bit-Quantization latancy(ms) |
|---|---|
| DoReFa | 1.227 |
| PACT | 1.224 |
| LQ-net | 1.198 |
| **Ours** | **0.584** |

## 5.1 Evaluation on the Robustness

Low-bit quantization can further improve the computation efficiency on the emergent DNN hardware accelerators[17]. As to practical application, we prefer to use the low-bit quantization method to compress the NN.However, model quantization will cause the accuracy loss, which has been shown in the Fig4. (a). The accuracies of both VGG16 and Squeezenet drop when quantized with less bits, where the phenomenon is more serious in Squeezenet . We can conclude that after the low-bit quantization, especially when the value is less than 5bit, the accuracy of CNN network decreases rapidly.

In order to further explore the influence of low-bit quantization on NN, we test the accuracy with the adversarial data to evaluate the robustness of the quantized NN, which is shown in the Fig.5 (b). The results also demonstrate that quantized Squeezenet is more susceptible to interference by adversarial attacks.

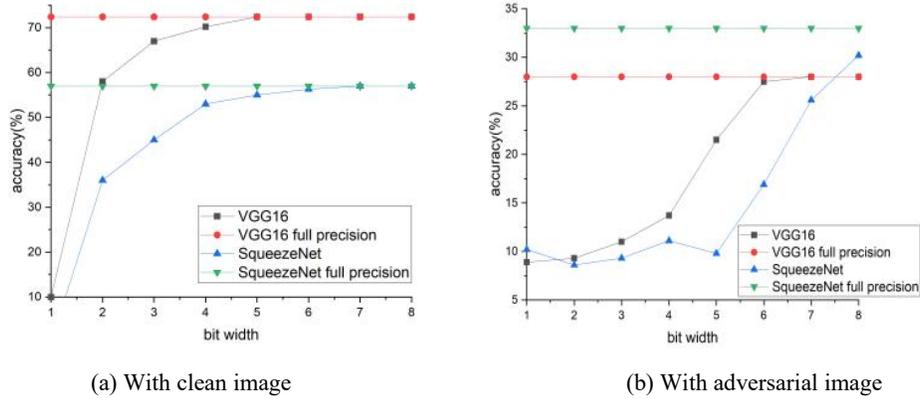

(a) With clean image　　　　　　　　　　　(b) With adversarial image
**Fig. 5.** Model accuracy with the presentation bit numbers

We evaluate the robustness using the ResNet20 by vanilla quantization method and our method with the adversarial samples, which generate from the CIFAR-10 test set by using FGSM attacker. The experimental results are shown in the Fig 5. Blue triangle represents the full-precision accuracy of ResNet 20 with the adversarial images.

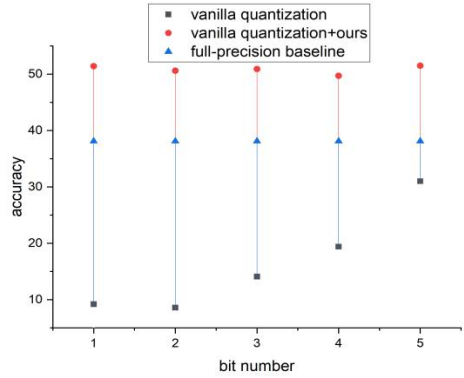

**Fig. 6.** Low-bit quantization accuracy of ResNet20 on CIFAR-10 with adversarial image

The robustness results are given in Fig 6., which demonstrates that our scheme with the structural loss and non-sensitive perturbation loss can preserve the robustness of LNN to some extent, and accuracies are improved by over 30%.

Furthermore, we also conducted experiments using PACT and our method to test the model convergence speed. Both methods use the same hyper-parameters. We illustrate the loss and accuracy curves in the first 50 epochs, which are shown in the Fig.7 (a) and (b), respectively.

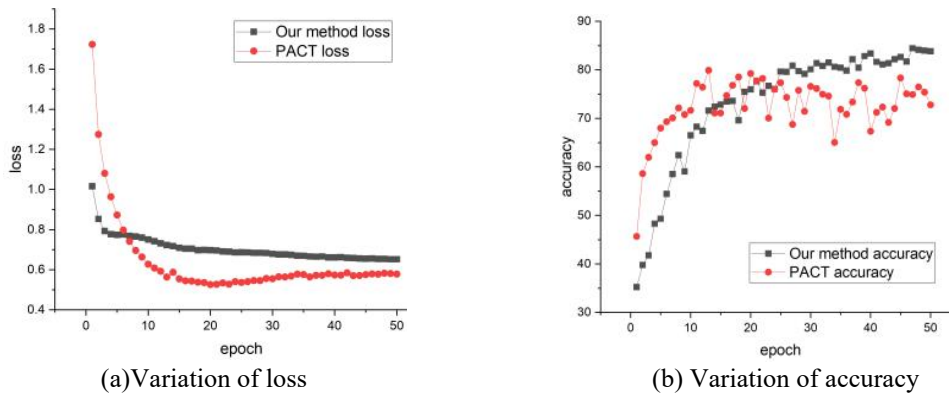

(a) Variation of loss　　　　　　　　　　　(b) Variation of accuracy
**Fig. 7.** Variation trend of loss and accuracy

It can be observed that during training, our loss reduces more steadily than PACT. At the same time, the accuracy obtained by our method increases smoothly and is better than PACT after 20 epochs. Thereby, the curves also verified the robustness of our algorithm. As a by-product, we can combine the other excellent quantization methods with our framework to further improve the performance as well as the robustness.

## 5.2 Overall Objective

Finally, during the implementation phase for the quantized NN network training, we synthesize the structural loss and non-sensitive perturbation loss together to guide the updating of quantized NN. The total loss is as follows：

$$L(total) = (1-\beta)*L(S) + \beta*L_\theta(G_\phi*) \qquad (11)$$

where $\beta$ represents the influence of structural information in the training process, which is tuned to be optimal through grid search. As shown in the Fig.8, we test various of low-bit quantization accuracy with different values of $\beta$, which vary from 0 to 1 with the step of 0.1.

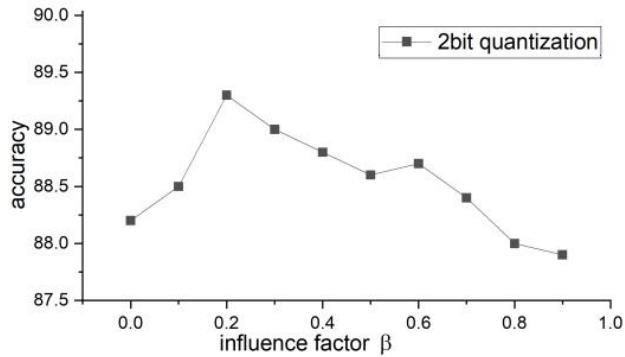

**Fig. 8.** Accuracy of ResNet20 on CIFAR-10 varies from the $\beta$

From Fig.7, we can see that the lower bit number is used, the smaller $\beta$ is needed to achieve the optimum solution, where we get the optimal 2-bit quantized NN at the value of 0.2, 0.3 and 0.5, respectively. If the value of $\beta$ is too small, it will cause the scarce participation degree of LS-GAN. On the other hand, a too large value of $\beta$ will lead to unstable prediction accuracy of quantized NN.

## 5.3 Ablation studies

**Structural-information loss.** In the experiment, we take 2-bit-precision and full-precision ResNet18 as the network of student and teacher, respectively. Results show that our proposed structural-information loss has better accuracy than others, results shown in the Table 4. The results of different settings for the student net are the averaged from three runs. From Table 1, we can see that the structural-information loss can improve the performance of student network, which verified that the structural information is crucial for the learning of the student network.

**Table 4.** The results of knowledge distillation training with different distillation loss.

| Student | Teacher | Distillation loss | Accuracy(%) |
|---|---|---|---|
| | | | 66.07 |
| | | Kullback-Leibler divergence loss | 66.68 |
| 2-bit-precision ResNet18 | full-precision ResNet18 | Knowledge distill loss | 66.65 |
| | | Jensen-Shannon divergence loss | 66.76 |
| | | Attention loss | 66.89 |
| | | Structural-Information loss($L_\theta(G_\phi*)$) | **67.10** |

**Self-reference quantization scheme.** The ablation studies for evaluating the self-reference quantization scheme are conduct on CIFAR-10 dataset with ResNet20. Table 5 shows the results of the full precision network and the quantized network using different variations of our method. When evaluating on the clean images, we can observe that structural information loss is the leading cause of accuracy improving, and the self-reference quantization scheme can improve the results further. Experiments conducted with adversarial image show that LS-GAN can improve the accuracy to some extent, but NPL is quite effective on the quantized model and even performs better than the full-precision model.

Table 5. The accuracy of ResNet20 in the proposed method with clean image and adversarial image. VQ = vanilla quantization method(4bit), SIL = structural information loss, NPL = Non-sensitive perturbation loss

| Method | Accuracy with clean image(%) | Accuracy with adversarial image(%) |
|---|---|---|
| ResNet20 | 91.8 | 38.4 |
| +VQ | 91.1 | 18.9 |
| +VQ+SIL | 91.5 | 20.5 |
| +VQ+SIL+GAN | 91.6 | 21.1 |
| +VQ+SIL+LS-GAN | 91.6 | 32.8 |
| +VQ+SIL+LS-GAN+NPL | 91.7 | 49.2 |

## 6  Conclusion

Considering the friendly nature of quantized NN for hardware, this paper intends to arise the attention on the quantization of NN and resolves the issues of performance degeneration and the declination of model robustness. In this paper, we propose a novel robustness-aware 2-bit quantization scheme to improve the performance and robustness, simultaneously. Specifically, the exploitation of structural information can lead to better classification results by preserving the representation capability of the quantized network and the non-sensitive perturbation loss function effectively guarantee the robustness and defensiveness of the quantized model. The experiments also verified that the proposed robustness-aware 2-bit quantization scheme also accelerates the convergence speed. Extensive experiments on CIFAR-10 and ImageNet datasets show that our method achieves the acceptable accuracy drops after 2bit quantization, and obtains almost no drops of performance on ResNet18 and ResNet20. In the meantime, the proposed method is also effective and robust under the FGSM adversarial samples attack, which is even better than the full precision counterparts.

## Declaration of Competing Interest

The authors declare that they have no known competing financial interests or personal relationships that could have appeared to influence the work reported in this paper.

## Acknowledgements

We are grateful to anonymous reviewers for their constructive comments. This work is partially supported by the National Science Foundation of China(NSFC) under Grant No. 61872017.

## References

[1] M. Courbariaux, I. Hubara, D. Soudry, et al.: Binarized Neural Networks: Training Deep Neural Networks with Weights and Activations Constrained to +1 or -1, 2016.
[2] M. Rastegari, V. Ordonez, J. Redmon, A. Farhadi.: Xnor-net: Imagenet classification using binary convolutional neural networks, In European Conference on Computer Vision, pages 525–542. Springer, 2016.

[3] Naveen Mellempudi, Abhisek Kundu, Dheevatsa Mudigere, Dipankar Das, Bharat Kaul, Pradeep Dubey.: Ternary Neural Networks with Fine-Grained Quantization, CoRR, abs/1705.01462, 2017.
[4] A. Zhou, A. Yao, Y. Guo, L. Xu, Y. Chen.: Incremental network quantization: Towards lossless cnns with low-precision weights, International Conference on Learning Representations, 2017.
[5] S. Zhou, Y. Wang, H. Wen, Q. He, Yuheng Zou.: Balanced Quantization: An Effective and Efficient Approach to Quantized Neural Networks, CoRR, abs/1706.07145, 2017.
[6] S. Zhou, Y. Wu, Z. Ni, X. Zhou, H. Wen, Y. Zou.: Dorefa-net: Training low bitwidth convolutional neural networks with low bitwidth gradients, arXiv preprint arXiv:1606.06160, 2016.
[7] J. Choi, Z. Wang, S. Venkataramani, P. I.-J. Chuang, V.: Srinivasan, K. Gopalakrishnan.: Pact: Parameterized clipping activation for quantized neural networks, arXiv preprint arXiv:1805.06085, 2018.
[8] Qi, Guo-Jun.: Loss-Sensitive Generative Adversarial Networks on Lipschitz Densities, International Journal of Computer Vision (2017): 1 - 23.
[9] A. Gholami, K. Kwon, B. Wu, Z. Tai, X. Yue, P. Jin, S. Zhao, K. Keutzer.: Squeezenext: Hardware-aware neural network design, Workshop paper in CVPR, 2018.
[10] F. Chollet.: Xception: Deep learning with depthwise separable convolutions, In Proceedings of the IEEE conference on computer vision and pattern recognition, 1251–1258, 2017.
[11] A. G. Howard, M. Zhu, B. Chen, D. Kalenichenko, W. Wang, T. Weyand, M. Andreetto, H. Adam.: Mobilenets: Efficient convolutional neural networks for mobile vision applications. arXiv preprint arXiv:1704.04861, 2017.
[12] N. Ma, X. Zhang, H.-T. Zheng, J. Sun.: Shufflenet v2: Practical guidelines for efficient cnn architecture design. In Proceedings of the European Conference on Computer Vision (ECCV), 116–131, 2018.
[13] Zoph B, Vasudevan V, Shlens J, et al.: Learning transferable architectures for scalable image recognition. Proceedings of the IEEE conference on computer vision and pattern recognition, 8697-8710, 2018.
[14] J. Wu, C. Leng, Y. Wang, Q. Hu, J. Cheng.: Quantized convolutional neural networks for mobile devices, In Proceedings of the IEEE Conference on Computer Vision and Pattern Recognition, 4820– 4828, 2016.
[15] J. Yim, D. Joo, J. Bae, J. Kim.: A gift from knowledge distillation: Fast optimization, network minimization and transfer learning, In Proceedings of the IEEE Conference on Computer Vision and Pattern Recognition, 4133–4141, 2017.
[16] S. Han, H. Mao, and W. J. Dally.: Deep compression: Compressing deep neural networks with pruning, trained quantization and huffman coding. International Conference on Learning Representations, 2016.
[17] K. Wang, Z. Liu, Y. Lin, J. Lin, and S. Han.: HAQ: Hardware-aware automated quantization, In Proceedings of the IEEE conference on computer vision and pattern recognition, 2019.
[18] T. Miyato, T. Kataoka, M. Koyama, et al.: Spectral Normalization for Generative Adversarial Networks, 2018.
[19] C. Zhu, S. Han, H. Mao, and W. J. Dally. Trained ternary quantization. In ICLR, 2017. 2.
[20] X. Lin, C. Zhao, and W. Pan. Towards accurate binary convolutional neural network. In NeurIPS. 2017. 2.
[21] Z. Cai, X. He, J. Sun, and N. Vasconcelos. Deep learning with low precision by half-wave gaussian quantization. In IEEE CVPR, 2017.
[22] D. Zhang, J. Yang, D. Ye, and G. Hua. Lq-nets: Learned quantization for highly accurate and compact deep neural networks. In ECCV, 2018.
[23] IR-NET
[24] Joseph Bethge, Marvin Bornstein, Adrian Loy, Haojin Yang, and Christoph Meinel. Training competitive binary neural networks from scratch. arXiv preprint arXiv:1812.01965, 2018
[25] Jian Guo, He He, Tong He, Leonard Lausen, Mu Li, Haibin Lin, Xingjian Shi, Chenguang Wang, Junyuan Xie, Sheng Zha, Aston Zhang, Hang Zhang, Zhi Zhang, Zhongyue Zhang, and Shuai Zheng. GluonCV and GluonNLP: Deep Learning in Computer Vision and Natural Language Processing. arXiv preprint arXiv:1907.04433, 2019.
[26] Y. Chen, C. Shen, X.-S. Wei, L. Liu, and J. Yang, "Adversarial PoseNet: A structure-aware convolutional network for human pose estimation," in Proc. IEEE Int. Conf. Comp. Vis., 2017, pp. 1212– 1221.
[27] P. Luc, C. Couprie, S. Chintala, and J. Verbeek, "Semantic segmentation using adversarial networks," arXiv: Comp. Res. Repository, vol. abs/1611.08408, 2016.
[28] K. Gwn Lore, K. Reddy, M. Giering, and E. A. Bernal, "Generative adversarial networks for depth map estimation from rgb video," in Proc. IEEE Conf. Comp. Vis. Patt. Recogn., 2018, pp. 1177–1185.
[29] Inoue H, Taylor R L. Laws of Large Numbers for Exchangeable Random Sets in Kuratowski-Mosco Sense[J]. Stochastic Analysis and Applications, 2006, 24(2):263-275.
[30] Gorshenin A , Korolev V , Zeifman A . Modeling Particle Size Distribution in Lunar Regolith via a Central Limit Theorem for Random Sums[J]. 2020.